\documentclass[11pt]{article}
\usepackage{acl-hlt2011}
\usepackage{times}
\usepackage{url}
\usepackage{latexsym}
\usepackage{multirow}
\usepackage{graphicx}
\usepackage{amssymb}
\usepackage{amsthm}
\usepackage{amsmath}
\usepackage{color}
\usepackage{balance}

\usepackage{algorithm,algorithmic}

\newcommand{\eat}[1]{\ignorespaces}

\title{A Universal Part-of-Speech Tagset}

\author{
  Slav Petrov\\
  Google Research\\
  New York, NY, USA\\
  {\tt slav@google.com}\\\And
Dipanjan Das\\
  Carnegie Mellon University\\
  Pittsburgh, PA, USA\\
  {\tt dipanjan@cs.cmu.edu}\\\And
  Ryan McDonald\\
  Google Research\\
  New York, NY, USA\\
  {\tt ryanmcd@google.com} 
}

\date{}
\begin{document}
\maketitle

\begin{abstract}
To facilitate future research in unsupervised induction of syntactic structure and 
to standardize best-practices, we propose a tagset that consists of twelve 
universal part-of-speech  categories. 
In addition to the tagset, we develop a mapping from 25 different
treebank tagsets to this universal set. 
As a result, when combined with the original treebank data, 
this universal tagset and mapping produce a dataset consisting of 
common parts-of-speech for 22 different languages. 
We highlight the use of this resource via two experiments, including
 one that reports competitive accuracies for unsupervised 
 grammar induction without gold standard part-of-speech tags.

\end{abstract}

\section{Introduction}

Part-of-speech (POS) tagging has received a great deal of attention as it is a critical
component of most natural language processing systems.
As supervised POS tagging accuracies for English (measured on 
the Wall Street Journal portion of the PennTreebank \cite{marcus.etal.93})
have converged to around 97.3\% \cite{toutanova.et.al.03,shen.et.al.09}, the attention has shifted to 
unsupervised approaches \cite{christodoulopoulos.et.al.10}.
In particular, there has been growing interest in both multi-lingual POS induction \cite{snyder.et.al.09b,naseem.et.al.09} and cross-lingual POS induction via treebank projection \cite{yarowsky.ngai.01,xi.hwa.05,das.petrov.11}.

Underlying these studies is the idea that a set of (coarse)
syntactic POS categories exist in similar forms across languages.
These categories are often called \emph{universals} to represent their cross-lingual
nature \cite{carnie.2002,newmayer.2005}.
For example, \newcite{naseem.et.al.09} used the Multext-East \cite{erjavec.et.al.04} corpus 
to evaluate their multi-lingual POS induction system, because it uses the same tagset for multiple languages. 
When corpora with common tagsets are unavailable, a standard approach is to manually define a mapping from
language and treebank specific fine-grained tagsets to a predefined universal set.
This was the approach taken by \newcite{das.petrov.11} to evaluate their cross-lingual POS projection system for six different languages.

To facilitate future research and to standardize best-practices, we propose a tagset that consists of twelve universal POS categories.
While there might be some controversy about what the exact tagset should be, 
we feel that these twelve categories cover the most frequent part-of-speech
that exist \eat{in one form or another} in most languages. 
In addition to the tagset, we also develop a mapping from fine-grained POS tags
for 25 different treebanks to this universal set. As a result, when combined with the original treebank data, this universal tagset and mapping produce a dataset consisting of common parts-of-speech for 22 different 
languages.\footnote{We include mappings for two different Chinese, German and Japanese treebanks.}
Both the tagset and mappings are made available
 for download at {\bf http://code.google.com/p/universal-pos-tags/}.

\begin{figure*}
\centering
\small
\setlength{\tabcolsep}{4pt}
\begin{tabular}{rllllllllllllll}
sentence: & The & oboist & Heinz & Holliger & has & taken & a & hard & line & about & the & problems & . \\
original: & \textsf{\textsc{Dt}} & \textsf{\textsc{Nn}} & \textsf{\textsc{Nnp}} & \textsf{\textsc{Nnp}} & \textsf{\textsc{Vbz}} & \textsf{\textsc{Vbn}} & \textsf{\textsc{Dt}} & \textsf{\textsc{Jj}} & \textsf{\textsc{Nn}} & \textsf{\textsc{In}} & \textsf{\textsc{Dt}} & \textsf{\textsc{Nns}} &. \\
universal:  & \textsf{\textsc{Det}} & \textsf{\textsc{Noun}} & \textsf{\textsc{Noun}} & \textsf{\textsc{Noun}} & \textsf{\textsc{Verb}} & \textsf{\textsc{Verb}} & \textsf{\textsc{Det}} & \textsf{\textsc{Adj}} & \textsf{\textsc{Noun}} & \textsf{\textsc{Adp}} & \textsf{\textsc{Det}} & \textsf{\textsc{Noun}} & \textsf{\textsc{.}} \\
\end{tabular}
\caption{Example English sentence with its language specific and corresponding universal POS tags.}
\label{fig:ex_mapping}
\end{figure*}

This resource serves multiple purposes. First, as mentioned previously, it is useful for building and evaluating unsupervised and cross-lingual taggers. Second, it also permits for a more reasonable comparison of accuracy across languages for supervised taggers. Statements of the form ``POS tagging for language X is harder than for language Y'' are vacuous when the tagsets used for the two languages are incomparable (not to mention of different cardinality).
Finally, it also permits language technology practitioners to train POS taggers with common tagsets across multiple languages. This in turn facilitates downstream application development as there is no need to maintain language specific rules due to 
differences in treebank annotation guidelines.

In this paper, we specifically highlight two use cases of this resource.
First, using our universal tagset and mapping, we run an experiment comparing POS tag accuracies for 25 different treebanks to evaluate POS tagging accuracy on a single tagset. Second, we combine the cross-lingual projection part-of-speech taggers of \newcite{das.petrov.11} with the grammar induction system of \newcite{naseem.etal.2010} -- which requires a universal tagset -- to produce a completely unsupervised grammar induction system for multiple languages, that does not require gold POS tags in the target language.

\section{Tagset}

While there might be some disagreement about the exact definition of 
an universal POS tagset \cite{myth}, it seems fairly indisputable that a set of coarse
POS categories (or syntactic universals) exists across all languages in one form or 
another \cite{carnie.2002,newmayer.2005}.
Rather than arguing over definitions, we took a pragmatic standpoint during the 
design of the universal POS tagset
and focused our attention on the POS categories that we expect to be most 
useful (and necessary) for users of POS taggers. 
In our opinion, these are NLP practitioners using
POS taggers in downstream applications, and NLP researchers using 
POS taggers in grammar induction and other experiments.

A high-level analysis of the tagsets underlying various treebanks
shows that the majority of tagsets are very fine-grained and language specific. 
In fact, \newcite{smith.eisner.2005} made a similar observation and defined
a collapsed set of 17 English POS tags (instead of the original 45) that has subsequently
been adopted by most unsupervised POS induction work.
Similarly, the organizers of the CoNLL shared tasks
on dependency parsing provide coarse (but still language specific) tags in
addition to the fine-grained tags used in the original treebanks \cite{conll.x,conll07}.
\newcite{mcdonald.07} identified eight different coarse POS tags when analyzing the errors of two dependency parsers on the 13 different languages from the CoNLL shared tasks.

Our universal POS tagset unifies this previous work and defines
the following twelve POS tags: 
\textsf{\textsc{Noun}} (nouns), \textsf{\textsc{Verb}} (verbs),
\textsf{\textsc{Adj}} (adjectives), \textsf{\textsc{Adv}} (adverbs),
\textsf{\textsc{Pron}} (pronouns), \textsf{\textsc{Det}} (determiners and articles),
\textsf{\textsc{Adp}} (prepositions and postpositions), \textsf{\textsc{Num}}
(numerals), \textsf{\textsc{Conj}} (conjunctions),
\textsf{\textsc{Prt}} (particles), \textsf{\textsc{`.'}} (punctuation marks)
and \textsf{\textsc{X}} (a catch-all for other categories such as abbreviations or foreign words). 

We did not rely on intrinsic definitions of the above categories.
Instead, each category is defined operationally.
For each treebank under consideration, we studied the exact POS tag definitions and
annotation guidelines and created a mapping from the original treebank tagset
to these universal POS tags. Most of the decisions were fairly clear.
For example, from the PennTreebank,
\textsf{\textsc{Vb}}, \textsf{\textsc{Vbd}}, \textsf{\textsc{Vbg}}, \textsf{\textsc{Vbn}},
\textsf{\textsc{Vbp}}, \textsf{\textsc{Vbz}} and \textsf{\textsc{Md}} (modal)
were all mapped to \textsf{\textsc{Verb}}.
A less clear case was the universal tag for particles, \textsf{\textsc{Prt}}, which
was mapped from \textsf{\textsc{Pos}} (possessive), \textsf{\textsc{Rp}} (particle) and \textsf{\textsc{To}} (the word `to'). In particular, the
\textsf{\textsc{To}} tag is ambiguous in the PennTreebank between infinitival markers
and the preposition `to'. Thus, under this mapping, some prepositions will be marked
as particles in the universal tagset. Figure \ref{fig:ex_mapping} gives an example
mapping for a sentence from the PennTreebank.

\begin{table*}[t]
	\centering
	\small
	\begin{tabular}{|l|l|c|c|c|c|}
		\hline
		\textbf{Language} & \textbf{Source} & \textbf{\# Tags} & \textbf{O/O} & \textbf{U/U} & \textbf{O/U} \\ \hline
		Arabic    & PADT/CoNLL07 \cite{PADT} & 21 & 96.1 & 96.9 & 97.0 \\
		Basque    & Basque3LB/CoNLL07 \cite{Basque} & 64 & 89.3 & 93.7 & 93.7 \\
		Bulgarian & BTB/CoNLL06 \cite{BTB} & 54 & 95.7 & 97.5 & 97.8 \\
		Catalan   & CESS-ECE/CoNLL07 \cite{CESS} & 54 &  98.5 & 98.2 & 98.8 \\
		Chinese & Penn ChineseTreebank 6.0 \cite{CTB} & 34 & 91.7 & 93.4 & 94.1 \\
		Chinese   & Sinica/CoNLL07 \cite{sinica} & 294 & 87.5 & 91.8 & 92.6 \\
		Czech     & PDT/CoNLL07 \cite{PDT} & 63 & 99.1& 99.1 & 99.1 \\
		Danish    & DDT/CoNLL06 \cite{DDT} & 25 & 96.2 & 96.4 & 96.9 \\
		Dutch     & Alpino/CoNLL06 \cite{Alpino} & 12 & 93.0 & 95.0 & 95.0 \\
		English & PennTreebank \cite{marcus.etal.93} & 45 & 96.7 & 96.8 & 97.7 \\
		French    & FrenchTreebank \cite{french.treebank} & 30 & 96.6 & 96.7 & 97.3 \\
		German    & Tiger/CoNLL06 \cite{Tiger} & 54 & 97.9 & 98.1 & 98.8 \\
		German  & Negra \cite{Negra} & 54 & 96.9 & 97.9 & 98.6 \\
		Greek     & GDT/CoNLL07 \cite{GDT} & 38 & 97.2 & 97.5 & 97.8 \\
		Hungarian & Szeged/CoNLL07 \cite{Szeged} & 43 & 94.5 & 95.6 & 95.8 \\
		Italian   & ISST/CoNLL07 \cite{ISST} & 28 & 94.9 & 95.8 &  95.8 \\
		Japanese & Verbmobil/CoNLL06 \cite{Verbmobil} & 80 & 98.3 & 98.0 & 99.1 \\
		Japanese & Kyoto4.0 \cite{Kyoto} & 42 & 97.4 & 98.7 & 99.3  \\
		Korean    & Sejong (http://www.sejong.or.kr) & 187 & 96.5 & 97.5 & 98.4 \\
		Portuguese & Floresta Sint\'{a}(c)tica/CoNLL06 \cite{Floresta} & 22 & 96.9 & 96.8 & 97.4 \\
		Russian    & SynTagRus-RNC \cite{SynTagRus} & 11 & 96.8 & 96.8 & 96.8 \\
		Slovene    & SDT/CoNLL06 \cite{SDT} & 29 & 94.7 & 94.6 & 95.3 \\
		Spanish    & Ancora-Cast3LB/CoNLL06 \cite{Cast3LB}& 47 & 96.3 & 96.3 & 96.9 \\
		Swedish    & Talbanken05/CoNLL06 \cite{Talbanken} & 41 & 93.6 & 94.7 & 95.1 \\
		Turkish    & METU-Sabanci/CoNLL07 \cite{MetuSabanci} & 31 & 87.5 & 89.1 & 90.2 \\ 
		\hline
	\end{tabular}
\caption{\label{table:posresults}Data sets, number of language specific tags in the
original treebank, and tagging accuracies for training/testing on 
the original (O) and the universal (U) tagset.
Where applicable, we indicate whether the data set was extracted from the
 CoNLL 2006 \cite{conll.x} or CoNLL 2007 \cite{conll07} versions of the corpora.
}
\end{table*}

Another case we had to consider is that some tag categories do not occur in all languages.
Consider for example the case of adjectives. While all languages
have a way of describing the properties of objects 
(which themselves are typically referred to with nouns), 
many have argued that Korean does not technically have adjectives,
but instead expresses properties of nouns via stative verbs \cite{kim.02}.
As a result, in our mapping for Korean, we mapped stative verbs to the universal
\textsf{\textsc{Adj}} tag.
In other cases this was clearer,
e.g. the Bulgarian treebank has no category for determiners or articles.
This is not to say that there are no determiners in the Bulgarian language.
However, since they are not annotated as such in the treebank, we are not able to include
them in our mapping.

The list of treebanks for which we have constructed mappings 
can be seen in Table \ref{table:posresults}.
One main objective in publicly releasing this resource 
is to provide treebank and language specific experts
a mechanism for refining these categories and the decisions we have made,
as well as adding new treebanks and languages.
This resource is therefore hosted as an open source project with 
version control.

\section{Experiments}

To demonstrate the efficacy of the proposed universal POS tagset,
we performed two sets of experiments.
First, to provide a language comparison, we trained the same supervised 
POS tagging model on all of the above treebanks and evaluated the tagging 
accuracy on the universal POS tagset.
Second, we used universal POS tags (automatically projected from English) as the 
starting point for unsupervised grammar induction, producing completely 
unsupervised parsers for several languages.

\subsection{Language Comparisons}

To compare POS tagging accuracies across different languages we trained
a supervised tagger based on a trigram Markov model \cite{brants.00} on all treebanks.
We chose this model for its fast speed and (close to) state-of-the-art accuracy 
without language specific tuning.\footnote{Trained on the English PennTreebank
this model achieves 96.7\% accuracy when evaluated on the original 45 POS tags.}

Table \ref{table:posresults} shows the results for all 25 treebanks when 
training/testing on the original (O) and universal (U) tagsets.
Overall, the variance on the universal tagset has been reduced by half
(5.1 instead of 10.4).
But of course there are still accuracy differences across the different languages.
On the one hand, given a golden segmentation, tagging Japanese is almost deterministic, resulting in a final accuracy of above 99\%.\footnote{Note that 
the accuracy on the universal POS tags for the two Japanese treebanks is almost the same.}
On the other hand, tagging Turkish,  an agglutinative language with
and average sentence length of 11.6 tokens, remains very challenging,
resulting in an accuracy of only 90.2\%.

It should be noted that the best results are obtained by training on the original
treebank categories and mapping the predictions to the 
universal POS tags at the end (O/U column). 
This is because the transition model based on the universal POS tagset
is less informative. 
An interesting experiment would be to train the latent variable tagger of 
\newcite{huang.eidelman.harper.09} on this tagset. Their model
automatically discovers refinements of the observed categories and
could potentially find a tighter fit to the data, than the one provided by the 
original, linguistically motivated treebank tags.

\subsection{Grammar Induction}

We further demonstrate the utility of the universal POS tags in a grammar induction experiment. 
To decouple the challenges of POS tagging and parsing, golden POS tags
are typically assumed in unsupervised grammar induction 
experiments \cite{carroll.charniak92,klein.manning.04}.\footnote{A less benevolent explanation for this
practice is that grammar induction from plain text is simply still too difficult.}
We propose to remove this unrealistic simplification by using POS tags 
automatically projected from English as the basis of a grammar induction model.

\newcite{das.petrov.11} describe a cross-lingual projection framework to learn
POS taggers without labeled data for the language of interest.
We use their automatically induced POS tags to induce syntactic dependencies. 
To this end, we chose the framework of \newcite{naseem.etal.2010}, 
in which a few universal syntactic rules (USR) are used to constrain a 
probabilistic Bayesian model.
These rules are specified using a set of universal syntactic categories, 
and lead to state-of-the-art grammar induction performance superior to previous methods, 
such as the dependency model with valence (DMV) \cite{klein.manning.04} 
and the phylogenetic grammar induction model (PGI) \cite{berg.klein.10}.

In their experiments, Naseem~et~al. also used a set of universal categories, 
however, with some differences to the tagset presented here.
Their tagset does not have punctuation and catch-all categories, 
but includes a category for auxiliaries. 
The auxiliary category helps define a syntactic rule that 
attaches verbs to an auxiliary head, which is beneficial for certain languages. 
However, since this rule is reversed for other languages, we omit it in our tagset. 
Additionally, they also used refined categories in the form of CoNLL treebank tags.
In our experiments, we did not make use of refined categories, 
as the POS tags induced by \newcite{das.petrov.11} were all coarse.

\begin{table}[t]
\centering
\begin{tabular}{|l|c|c||c||c|}
  \hline
\textbf{Language} & \textbf{DMV} & \textbf{PGI} & \textbf{USR-G} & \textbf{USR-I}\\
\hline
Danish & 33.5 & 41.6 & 55.1 & 41.7 \\
Dutch   & 37.1 & 45.1 & 44.0 & 38.8 \\
German   & 35.7 & -\footnotemark \addtocounter{footnote}{-1} & 60.0 & 55.1 \\
Greek & 39.9 & -\footnotemark \addtocounter{footnote}{-1} & 60.3 & 53.4 \\
Italian & 41.1 & -\footnotemark & 47.9 & 41.4 \\
Portuguese   & 38.5 & 63.0 & 70.9 & 66.4 \\
Spanish   & 28.0 & 58.4  & 68.3 & 43.3\\
Swedish   & 45.3 & 58.3 & 52.6 & 59.4 \\
\hline
\end{tabular}
\caption{\label{table:giresults} 
Grammar induction results in terms of directed dependency accuracy.
{DMV}, {PGI} and use fine-grained gold POS tags,
while {USR-G} and {USR-I} uses gold and automatically projected universal POS tags respectively.
}
\end{table}
\footnotetext{Not reported by \newcite{berg.klein.10}.}

We present results on the same eight Indo-European languages as \newcite{das.petrov.11},
so that we can make use of their automatically projected POS tags.
For all languages, we used the treebanks released as a part of the CoNLL-X \cite{conll.x} 
shared task. We only considered sentences of length 10 or less, after the removal of punctuations. 
We performed Bayesian inference on the whole treebank and report dependency attachment accuracy.

Table~\ref{table:giresults} shows directed dependency accuracies for the DMV and PGI
models using fine-grained gold POS tags.
For the USR model, it reports results on gold universal POS tags (USR-G) 
and automatically induced universal POS tags (USR-I).
The USR-I model falls short of the USR-G model, but
has the advantage that it does not require any labeled data from the target language.
Quite impressively, it does better than DMV for all languages, 
and is competitive with PGI, 
even though those models have access to fine-grained gold POS tags.

\section{Conclusions}

We proposed a POS tagset consisting of twelve categories that exists across languages
and developed a mapping from 25 language specific tagsets to this universal set.
We demonstrated experimentally that the universal POS categories generalize well 
across language boundaries on an unsupervised grammar induction task, giving
competitive parsing accuracies without relying on gold POS tags. The tagset 
and mappings are available for download at
{\bf http://code.google.com/p/universal-pos-tags/}

\section*{Acknowledgements}
We would like to thank Joakim Nivre for allowing us to use a preliminary tagset mapping used in the work of \newcite{mcdonald.07}. The second author was supported in part by NSF grant IIS-0844507.


\balance
\bibliographystyle{acl}
\bibliography{bib}

\end{document}